%% file: main.tex
\documentclass[10pt,twocolumn,letterpaper]{article}

\usepackage[pagenumbers]{cvpr} 

\input{preamble}

\definecolor{cvprblue}{rgb}{0.21,0.49,0.74}
\usepackage[pagebackref,breaklinks,colorlinks,allcolors=cvprblue]{hyperref}

\def\paperID{4243} 
\def\confName{CVPR}
\def\confYear{2026}

\usepackage[dvipsnames]{xcolor}

\title{FROMAT: Multiview Material Appearance Transfer via \\Few-Shot Self-Attention Adaptation}

\author{Hubert Kompanowski\\
Trinity College Dublin\\
\and
Varun Jampani\\
Arcade AI\\
\and
Aaryaman Vasishta\\
AMD\\
\and
Binh-Son Hua\\
Trinity College Dublin\\
}

\begin{document}

\twocolumn[{%
\renewcommand\twocolumn[1][]{#1}%
\maketitle
\vspace{-25pt}
\begin{center}
   \includegraphics[width=\linewidth]{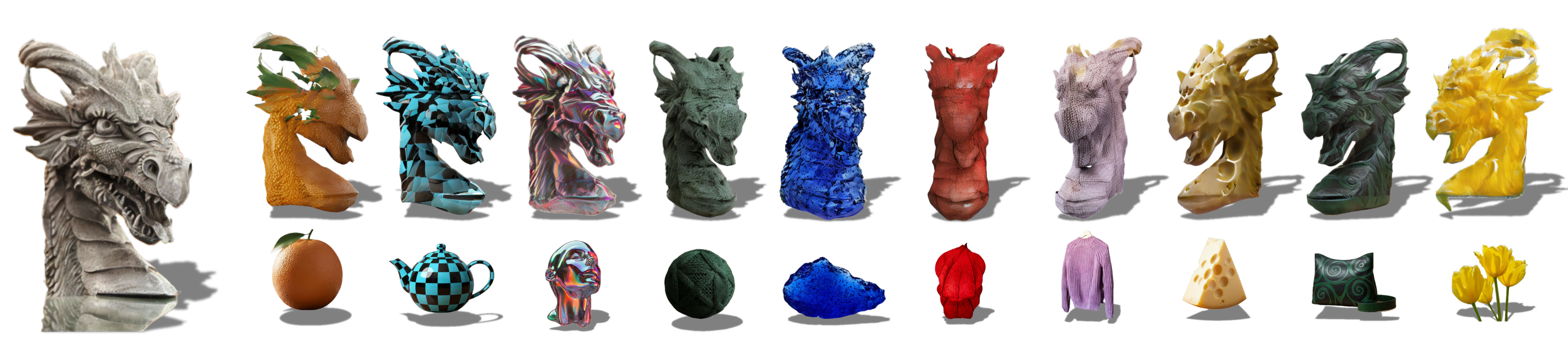}
    \captionof{figure}{Multiview material appearance transfer from a single input image (leftmost, the dragon head) and a reference appearance image. Please see the videos for all views.}
    \label{fig:teaser}
\end{center}
}]

\input{sec/0_abstract}

\input{sec/1_intro}
\input{sec/2_related_works}
\input{sec/3_method}

\input{sec/4_results}

\input{sec/5_conclusion}

{
    \small
    \bibliographystyle{ieeenat_fullname}
    \bibliography{main}
}

\input{sec/X_suppl}

\end{document}

%% file: preamble.tex
\usepackage{microtype}



%% file: sec/0_abstract.tex
\begin{abstract}

Multiview diffusion models have rapidly emerged as a powerful tool for content creation with spatial consistency across viewpoints, offering rich visual realism without requiring explicit geometry and appearance representation. However, compared to meshes or radiance fields, existing multiview diffusion models offer limited appearance manipulation, particularly in terms of material, texture, or style. 

In this paper, we present a lightweight adaptation technique for appearance transfer in  multiview diffusion models. 
Our method learns to combine object identity from an input image with appearance cues rendered in a separate reference image, producing multi-view-consistent output that reflects the desired materials, textures, or styles. This allows explicit specification of appearance parameters at generation time while preserving the underlying object geometry and view coherence. 
We leverage three diffusion denoising processes responsible for generating the original object, the reference, and the target images, and perform reverse sampling to aggregate a small subset of layer-wise self-attention features from the object and the reference to influence the target generation. 
Our method requires only a few training examples to introduce appearance awareness to pretrained multiview models. The experiments show that our method provides a simple yet effective way toward multiview generation with diverse appearance, advocating the adoption of implicit generative 3D representations in practice. 

\end{abstract}

%% file: sec/1_intro.tex
\section{Introduction}
\label{sec:intro}

The paradigm for 3D content creation is undergoing a significant shift. Traditional pipelines, reliant on explicit 3D representations like meshes and textures, are increasingly being complemented by implicit and view-based approaches. Recent generative models, particularly those based on diffusion, have demonstrated a remarkable ability to produce high-fidelity, 3D-consistent multiview image sets or videos directly from text or a single image \cite{shi2024MVDream, liu2024SyncDreamer, Wonder3D,li2024era3d, voleti2024sv3d,zhou2025seva}. These methods leverage large pre-trained image or video generative models to establish multiview imagery as a viable, self-contained representation for 3D objects or scenes. This trend aligns with a growing interest in generative models that represent 3D structure and motion implicitly, such as in video models or world models, rather than through explicit geometry.

Despite this rapid progress, a critical limitation persists. For traditional explicit 3D representations like 3D meshes, there exists well developed tools for appearance control such as Blender, Houdini, 3DS Max. 
By contrast, editing implicit 3D data like multiview images remain an underexplored problem.   
For example, a user cannot simply instruct a multiview model to change the object appearance to a shiny metal material, or generate the object by following a specific texture pattern from a reference image. 
Existing multiview diffusion pipelines treat an object appearance (its material or texture) as an inseparable part of its identity, which makes on-demand editing limited. 

A closely relevant line of research to appearance editing is general style transfer~\cite{hertz2024stylealigned,jeong2024vsp,jeong2025stylekeeper} and object material editing~\cite{cheng2024zest,garifullin2025materialfusion}. 

However, these methods are based on either generative models for single views~\cite{cheng2024zest,garifullin2025materialfusion}, which do not guarantee consistent results across multiple views, or tightly coupled to traditional 3D mesh pipeline~\cite{huang2024materialanything}. 
To our knowledge, no existing method offers direct surface appearance manipulation for a generative multiview diffusion model yet.

In this paper, we introduce an appearance transfer mechanism for multiview diffusion models via a lightweight, model-agnostic adaptation framework. 
We leverage image-to-multiview diffusion models~\cite{li2024era3d,voleti2024sv3d,zhou2025seva} and formulate material appearance transfer using a three-stream denoising framework. 
Our method aims to generate multiview images in the main stream so it inherits the object identity from an input image and the appearance specified in a reference image (see Fig.~\ref{fig:arch} for an illustration of this concept). 
Our adaptation operates by integrating key-value features from the object identity stream and reference appearance stream to the main stream via a learnable attention mixing mechanism.  
We demonstrate the importance of optimizing the attention mixing for each self-attention layer in the main stream, and the effectiveness of appearance transfer across material types. 

Our attention mixing can be trained in a few-shot regime, requiring as little as one object-reference pair to introduce the capability of appearance transfer in multiview diffusion model. 

After adaptation, the model can generalize to perform material appearance transfer on novel objects with new materials, producing view-consistent multiview images that preserve both the object identity and the reference appearance in details. 
We show that the adapted model accepts flexible appearance inputs, including physically-based rendered materials, texture patterns, real-world exemplar photographs while offering view consistencies. 

Our contributions are as follows.
\begin{itemize}
    \item A model-agnostic adaptation framework that introduces explicit, few-shot appearance transfer to pre-trained multiview diffusion models based on self-attention transfer across three denoising processes;
    \item A highly efficient gradient descent training strategy that adapts a pretrained multiview model in under 30 minutes on a single consumer GPU by freezing all base weights and training only new attention-mixing parameters. There is no manual parameter selection or grid search required;
    \item An extensive set of experiments demonstrating state-of-the-art results on material appearance transfer
    while maintaining geometric and multiview consistency.
\end{itemize}

%% file: sec/2_related_works.tex
\section{Related Works}
\label{sec:related}

\paragraph{Image Editing.}
Our work takes inspirations from state-of-the-art image generation and editing using diffusion models~\cite{rombach2022high,zhang2023controlnet,brooks2022instructpix2pix,kawar2023imagic,xie2024sana,xie2025sana}, particularly style transfer methods~\cite{hertz2024stylealigned,jeong2024vsp,chung2024styleinjection,alaluf2024crossattn,wang2024instantstyle,jeong2025stylekeeper}. 
In diffusion models, it is well known that image features interact to one another via self-attention layers. 
This interaction can be leveraged for style transfer by letting the target image attend to itself and reference image features~\cite{hertz2024stylealigned}, or inject/swap its key-value pairs using those of the reference image~\cite{jeong2024vsp,chung2024styleinjection,alaluf2024crossattn,wang2024instantstyle,jeong2025stylekeeper}.
Our work introduces self-attention swapping to the multiview setting and the material appearance transfer task. 
Unlike previous methods that are based on zero-shot transfer via manual attention layers selection, we use few-shot optimization~\cite{sohn2023styledrop,ruiz2022dreambooth} to identify optimal self-attention swapping configuration at specific layers, thereby introducing material awareness into multiview appearance transfer. 

While conventional style transfer methods address general appearance adaptation, there exists dedicated methods aiming to capture and transfer particular visual attributes such as material or texture.
Zero-shot material transfer methods like ZeST~\cite{cheng2024zest} and MaterialFusion~\cite{garifullin2025materialfusion} utilizes IP-Adapter to encode material features from a reference image together with Diffusion Inpainting, Depth Estimator or Guide-and-Rescale~\cite{gar} for addressing the object shape and localization control. Marble~\cite{cheng2025marble}, built on top of ZeST, introduces additional trainable lightweight MLP that predicts material editing direction in the CLIP-space, leading to improved material attribute control.

In addition to zero-shot transfer methods, there also exists finetuning-based techniques for appearance editing, e.g.,  DreamBooth~\cite{ruiz2022dreambooth} for artistic or personalized image generation, Alchemist~\cite{sharma2024alchemist} for parametric control of material properties like albedo, metal, roughness, transparency.

Compared to these methods, our method only trains on a tiny set of parameters while being able to generalize to new appearances.
Additionally, a special class of image editing methods opt to operate in the intrinsic-image latent space~\cite{zeng2024rgb,lyu2025intrinsicedit,lopes2025matswap}, aiming for precise manipulation of scene appearance including lighting and shadows. 
In contrast to these methods, our method operates on an image-to-multiview diffusion model and utilizes its existing capability of understanding different semantic concepts of the generated content for material appearance transfer. We address the transfer problem via a generic self-attention swapping framework and few-shot adaptation on synthetic data, removing the need of additional models to handle appearance extraction and targeted application separately.

\paragraph{3D Editing.}

Recent advances in 3D representations such as neural radiance fields (NeRFs)~\cite{mildenhall2020nerf} and 3D Gaussian splatting~\cite{kerbl3Dgaussians} also lead to new capabilities for appearance generation and editing~\cite{chen2025advances}. 
Inspired by image editing methods, a natural approach for 3D appearance editing is to edit view images and then update the underlying 3D representations~\cite{haque2023instructnerf,shum2024fusion}.
One can also separate appearance from geometry representation in a radiance field and learn to update the appearance features with styles~\cite{pang2023locally,liu2024stylegaussian}. 
There also exists a view-dependent reconstruction for NeRFs and 3D Gaussians to improve reflective and material-aware rendering~\cite{verbin2022refnerf,zhang2025refgs,zhang2025materialrefgs,ye2024gsdr,jiang2024gaussianshader,verbin2024nerfcasting}.

In the area of text-to-3D generation, score distillation techniques~\cite{poole2022dreamfusion} can be repurposed to distill a pretrained text-to-image model for learning to generate materials~\cite{chen2023fantasia3d,zhang2024dreammat,huang2024materialanything} and textures~\cite{chen2023text2tex,youwang2024paintit,huo2024texgen,deng2024flashtex}. 

These materials can then be applied to 3D assets for rendering using traditional computer graphics pipelines. 
Since material representation is essential for the appearance of the 3D surface, material reconstruction methods~\cite{guo2020MaterialGAN,hu2022material,zhou2022tilegen,vainer2024pbr,vecchio2024controlmat,lopes2024materialpalette,ma2025picker} focus on improving the quality of visual appearance by predicting fine-grained, high-resolution, physically plausible material from casually captured photographs. 

Compared to these 3D appearance modeling methods, our work embraces a new paradigm that leverages multiview images for implicit 3D representation, directly performing image-based appearance control on multiview data, and eliminating the need for rendering from triangle meshes or radiance fields.

\paragraph{Multiview and Video Editing}
Pretrained text-to-image diffusion models can be modified to include image prompts using a lightweight adapter such as IP-Adapter~\cite{ye2023ipadapter}, facilitating  image editing tasks. This technique was extended to learn image-to-multiview generation~\cite{huang2025mvadapter}, but these adapters still require expensive training, limiting their applicability to material appearance transfer. 
Among text-to-image(s) generation models, 
multiview generation~\cite{shi2024MVDream,li2024era3d,liu2024SyncDreamer,gao2024cat3d,voleti2024sv3d,zhou2025seva} demonstrated great promise in benefiting from the best of image and 3D generation. 
These generative models share similar network architecture to text-to-image models while being able to produce novel views with high-quality view consistency.
However, appearance control for multiview models remains underexplored, which motivates our work. 
Recently, video generation models have also been adopted for multiview generation and editing due to their ability to generate consistent video frames~\cite{zuo2024videomv}. 
In particular, SViM3D~\cite{engelhardt2025svim3d} predicts material parameters and surface normals from an input image using a pretrained video diffusion model~\cite{voleti2024sv3d}, the results of which could be used for reconstructing the underlying 3D assets and for rendering based on physical properties. 
Following this line of work, our work is grounded in the concept of multiview generation, and our method can be adapted to various image-to-multiview models.

%% file: sec/3_method.tex
\section{Method}
\label{sec:method}

\begin{figure*}[t]
    \centering
    \includegraphics[width=0.8\linewidth]{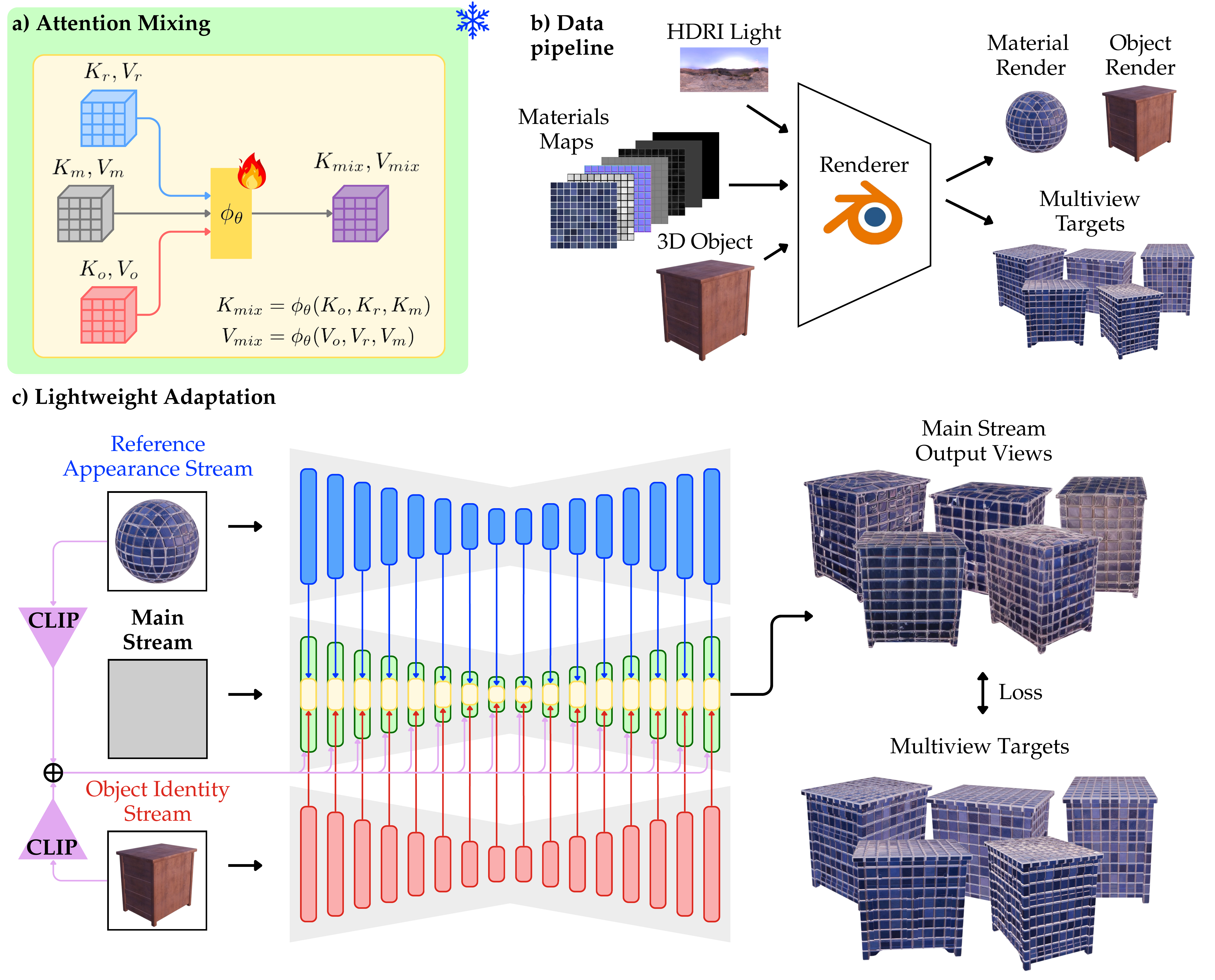}
    \caption{Overview of our method. a) Attention mixing mechanism. b) Training data generation pipeline. c) Three-stream denoising framework. We introduce Attention Mixing, each with its own mixing weights, for every self-attention block in the main stream denoising network. The mixing weights are optimized based on a few 3D objects rendered to multiview images. At inference, the main stream can then perform multiview appearance transfer from arbitrary pairs of object-reference images. }
    \label{fig:arch}
\end{figure*}

\paragraph{Overview.} 
Our method builds upon image-to-multiview diffusion models~\cite{li2024era3d, zhou2025seva, voleti2024sv3d} and adapts their reverse sampling processes to enable appearance control. 
The core idea is to disentangle geometry and appearance during image generation and perform appearance control based on three denoising processes, each being responsible for object identity, reference appearance, and target output. 
For brevity, we refer to these denoising processes as the object identity stream, the appearance reference stream, and the target (main) stream. 
We then introduce a minimal set of new trainable parameters based on the model's self-attention layers, enabling attention mixing that transfers the object's shape and the reference material into the main stream.
Finally, we train this attention mixing mechanism on a small set of synthetic data samples to activate view-consistent appearance control. 
A diagram of our method is shown in~\Cref{fig:arch}.
We detail our method in the following section.

\subsection{Approach}

We modify the denoising U-Net of the diffusion model to operate with three parallel-but-connected streams, as illustrated in \Cref{fig:arch}. All streams share the frozen weights of the original U-Net, but process different inputs and have distinct roles.

\begin{itemize}
    \item \textbf{Object identity stream:} Conditioned on the object input image, this stream performs standard, unmodified image to multiview generation. It serves to preserve the geometric identity and structural integrity of the object.
    
    \item \textbf{Appearance reference stream:} This stream is conditioned on the appearance reference image (e.g., a material sphere). Its purpose is to extract and propagate the essential appearance features: texture, color, and reflectance.
    
    \item \textbf{Main stream:} This is the primary synthesis stream that generates the final output. It composes its features from the other two streams via our proposed attention mixing mechanism, resulting in novel views of the target object with the specified appearance.
\end{itemize}

Let us now introduce our attention-mixing mechanism, which occurs on the main stream. 
In each layer of self-attention, let us denote the key, query, and value tensors of the object, appearance, and main streams as $(Q_{o}, K_{o}, V_{o})$, $(Q_{r}, K_{r}, V_{r})$, and $(Q_m, K_m, V_m)$, respectively.
The main stream computes its attention output using its own queries $Q_m$, but with a mixed set of keys and values defined by: 
\begin{align}
    K_{mix} &= \phi_\theta( K_o, K_r, K_m ) \label{eq:k_mix} \\
    V_{mix} &= \phi_\theta( V_o, V_r, V_m ) \label{eq:v_mix}
\end{align}
where $\phi$ is the aggregator function with a set of learnable parameters $\theta$. 
There are many possibilities to design this aggregator function. 
Here we aim for an aggregator with small number of parameters so that it can be easily learned from a small dataset. 
A simple definition of the aggregator function is therefore the linear function:
\begin{align}
    \phi_\theta(x_o, x_r, x_m) = \alpha_o x_o + \alpha_r x_r + \alpha_m x_m 
\end{align}
where $\theta = (\alpha_o, \alpha_r, \alpha_m)$ are learnable, view-dependent mixing weights for that specific attention layer. 
We enforce $\sum \alpha_i = 1$ by computing the weights from a small learnable parameter vector passed through a softmax function.
The final attention output for the main stream is then computed as $\mathrm{attention}(Q_m, K_{mix}, V_{mix})$.

Our per-layer attention mixing is inspired by \cite{hertz2024stylealigned,jeong2024vsp}, which shows that different attention layers capture features at different semantic levels. 
However, previous methods are based on manual selection of attention layers with fixed mixing weights. 
Our attention mixing generalizes this technique with a trainable scheme, which allows our model to learn, for example, to pull low-level texture features from the reference stream (e.g., at finer layers) while taking structural features from the object stream (e.g., at coarser layers) to synthesize a coherent final result.

\subsection{Training Specification}
To make our attention mixing work in practice, it is critical to learn the aggregator function for appearance transfer on specific data samples. 
As the number of parameters of our aggregator per layer is relatively small (e.g., 48 parameters for each self-attention layer), we initially opt for simulated annealing based optimization techniques to find these parameters. 
However, we found that this optimization process is very slow to converge. 
Therefore, we propose to use gradient descent optimization to perform the training. 
We observe that to successfully adapt the model, it is critical to adhere to the design of the image-to-multiview architecture and training details including image conditioning, diffusion loss function and noise schedule when the model is (pre-)trained. 
Given this requirement, we provide the training details for the following image-to-multiview models, where the training strategy is publicly available or can be easily reproduced. 
In particular, our implementation is built on recent multiview diffusion models such as Era3D~\cite{li2024era3d} and Stable Virtual Camera~\cite{zhou2025seva}. 

\noindent\textbf{Era3D.} This is an image-to-multiview model with the architecture extended from Stable Diffusion 2.1~\cite{rombach2022high} (image resolution 512x512). Era3D~\cite{li2024era3d} leverages a canonical camera setting (similar to the rectified epipolar camera setting with zero elevation camera angles) so that attention across frames can be applied in a row-wise manner. 

\noindent\textbf{Stable Virtual Camera (SEVA).} This is the state-of-the-art novel view synthesis model built upon the concept of multiview diffusion (image resolution 576x576). 
SEVA~\cite{zhou2025seva} is extended from Stable Diffusion 2.1 but includes major changes including scale to 1.3B parameters and training on video data. More importantly, SEVA can render new camera views on an arbitrary trajectory, making it a powerful technique for implicit 3D modeling. 
We predict 21 novel views from the input. 

By default, we will discuss implementation details based on SEVA. Details based on Era3D are available in the supplementary material.

\paragraph{Input Frame Conditioning.}
A common issue of appearance transfer in a multiview diffusion model is that the input frame, e.g., very often the frame showing the front of the object, is used as the condition and has a strong influence on the output images, making attention transfer ineffective. 
For example, the transfer tends to keep the original input frame in the output without any appearance changes. 
To address this issue, we remove concatenation-based conditioning from main stream and rely on transferring information from object identity stream and modified cross-attention (CLIP) conditioning to produce the output. 
Additionally, for SEVA, we introduce one more frame that follows the input image camera position, but is set to being a target instead of input and thus is being generated during de-noising process. This allows the transfer to occur on newly introduced frame and the input image frame is being discarded after the generation is completed.

For Era3D removing conditioning from main stream was sufficient to deal with the strict alignment between predicted frame and input image. This is because Era3D additionally predicts normal maps, which, when properly optimized, helps with the input frame issue, by allowing stronger reference mixing without compromising the object structure. Details are presented in the supplementary material.

\paragraph{Data and Training Pipeline.}
To perform training, we prepare a triplet of images for each sample: (1) an image of the object, (2) a reference image of the appearance (e.g., a rendered material sphere), and (3) a set of ground truth multiview images of the object with the target appearance. 

During training, we freeze all original weights of the pre-trained image-to-multiview diffusion model, and only train the attention mixing weights $(\alpha_{o}, \alpha_{r}, \alpha_{m})$ for each adapted self-attention layer. We optimize the model using the original diffusion objective and noise schedule applied to the output of the main stream. 
We perform our training on a state-of-the-art consumer GPU, e.g. NVIDIA RTX 5090, and it takes less than 30 minutes for the adaptation to converge.

\paragraph{Per-layer Optimization.} 

Our approach introduces learnable mixing weights for each self-attention layer, and optionally per input frame, enabling the model to selectively integrate information from the reference and identity streams. This optimization replaces manual layer selection—which is model-dependent and difficult to tune for new architectures—and allows the system to automatically identify the most informative layers for appearance control. Extending the mixing to operate per frame further increases flexibility, enabling the model to handle challenging viewpoints independently and improving overall robustness. To reduce interference between streams, we apply an argmax selection during inference, choosing the dominant stream per layer and per frame instead of relying on soft activations; this leads to sharper details and more stable appearance transfer. We additionally remove concatenation-based conditioning from the main stream to prevent over-alignment with the input frame, while retaining cross-attention (CLIP) conditioning using the averaged CLIP embeddings of the object and reference images. Finally, increasing the CFG scale of the main stream (from 1.2–2.0 to 5.0–7.0) amplifies the influence of the appearance and identity streams, improving both global structure and material fidelity. These design choices and their impact are analyzed in detail in our ablation study.

%% file: sec/4_results.tex
\section{Experimental Results}
\label{sec:results}

\subsection{Evaluation Overview}

\paragraph{Baselines.}
To ensure fair comparisons, we construct baseline methods that perform appearance transfer in a multiview setting. 
We adopt the following approach: we first apply a 2D material transfer method to an input image and a reference material image, and then lift the output image to multiple views using a pretrained image-to-multiview model.
We use Marble~\cite{cheng2025marble}, MaterialFusion~\cite{garifullin2025materialfusion}, and ZeST~\cite{cheng2024zest} as our material transfer methods.
By default, we use the Stable Virtual Camera (SEVA)~\cite{zhou2025seva} as the pretrained image-to-multiview model. 
Additional comparisons with other models (e.g. Era3D) are provided in the supplementary material.

\paragraph{Data.}
We construct our evaluation dataset by combining materials from the \textit{MatSynth} dataset~\cite{vecchio2024matsynth}, 3D objects from \textit{Objaverse}~\cite{deitke2023objaverse}, and environment lighting from the \textit{PolyHaven HDRI} library. The \textit{MatSynth} test split provides a diverse collection of materials designed to cover a wide range of reflectance, translucency, and texture properties.  

For each material, we apply it to distinct 3D objects sourced from \textit{Objaverse} and render the resulting scenes under varying illumination conditions using Blender Cycles. We employ HDR environment maps from \textit{PolyHaven} for natural lighting, as well as a standard three-point light setup for controlled comparisons. The camera trajectories follow an orbital path around the object, with elevation angles uniformly sampled from $[0^\circ, 30^\circ]$ and azimuthal rotations from $[0^\circ, 90^\circ, 180^\circ, 270^\circ]$ for quantitative evaluation.  

For additional qualitative experiments, we select a diverse subset of materials and objects that are not present in the quantitative set. We further include \textit{real life} examples, as well as generated images obtained using the \textit{Nano Banana} model~\cite{Fortin2025NanoBanana}. These examples allow us to assess robustness under more varied and less controlled conditions.

\subsection{Quantitative Comparison.}

We provide a quantitative comparison with the baseline methods in \Cref{tab:test_results}. Our method demonstrates superior performance across all metrics, achieving a significant improvement in PSNR and CLIP-i, indicating greater fidelity in both reconstruction and semantic alignment with the material.

\begin{table}[t]

\caption{Quantitative comparison for material application from a single-view reference. Our method significantly outperforms baselines in reconstruction quality (PSNR, LPIPS, FID) and material-shape semantic alignment (CLIP-i).}
\label{tab:test_results}

\centering
\begin{tabular}{lcccc}
\toprule
Base Method & PSNR $\uparrow$ & CLIP-i $\uparrow$ & LPIPS $\downarrow$ & FID $\downarrow$  \\
\midrule
ZeST  & 16.72 &  0.885 & 0.145 & 74.66 \\
MaterialFusion   & 16.71 &  0.841 & 0.156 & 96.80\\
MARBLE   & 17.15 &  0.881 & 0.139 & 75.16 \\

\textbf{Ours}  & \textbf{18.74}  & \textbf{0.907}& \textbf{0.119}& \textbf{61.80} \\

\bottomrule
\end{tabular}

\end{table}

\subsection{Qualitative Comparison.}

\begin{figure*}[t!]
    \centering
    \includegraphics[width=\linewidth]{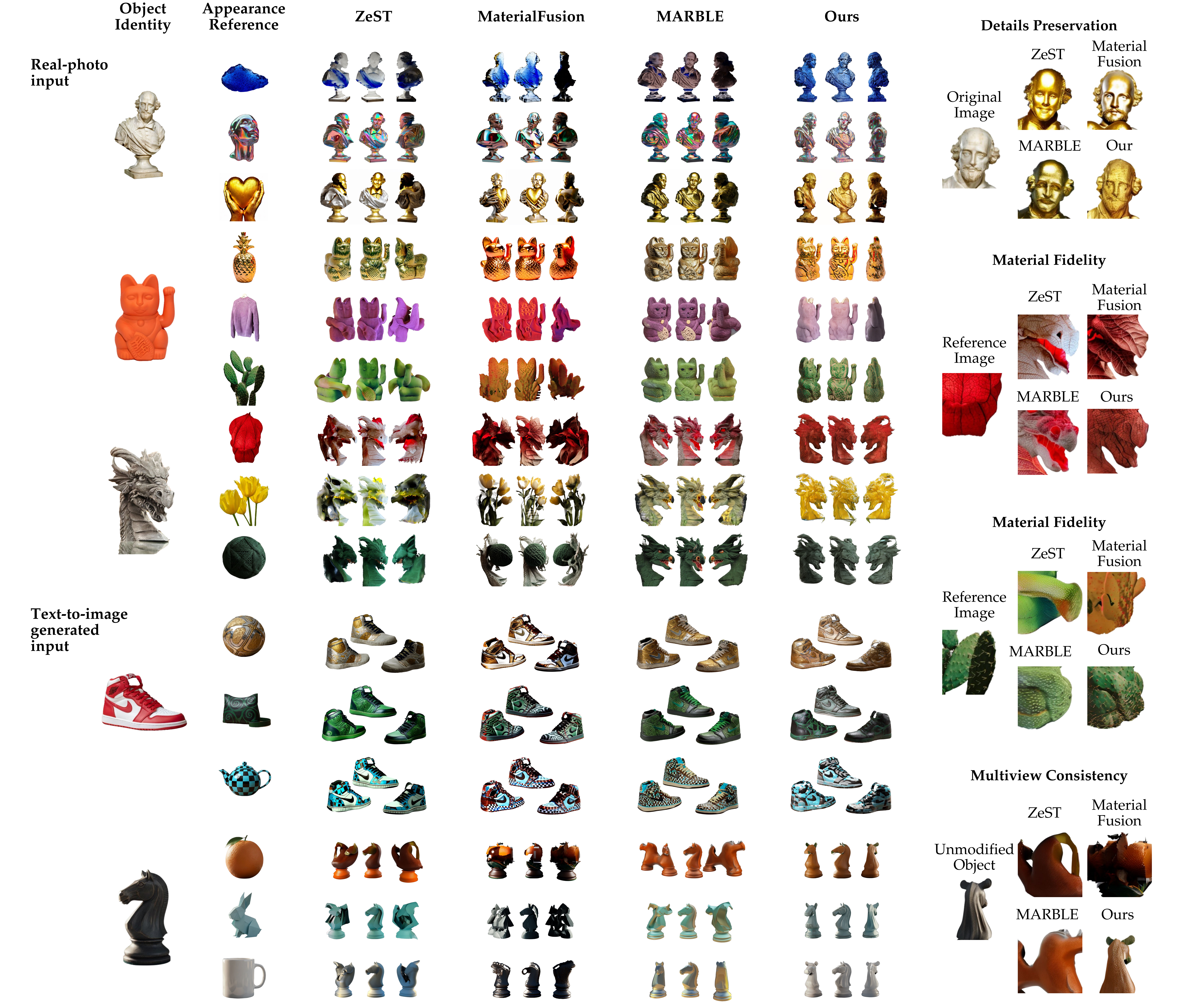}
    \caption{Qualitative comparison of our method with baseline on real photo inputs and generated images inputs. Our method successfully preserves object identity and achieves plausible material application, while the baselines struggle with lifting modified input image to multiview.}
    \label{fig:ours_vs_baselines}
\end{figure*}

\begin{figure}[ht]
    \centering
    \includegraphics[width=\linewidth]{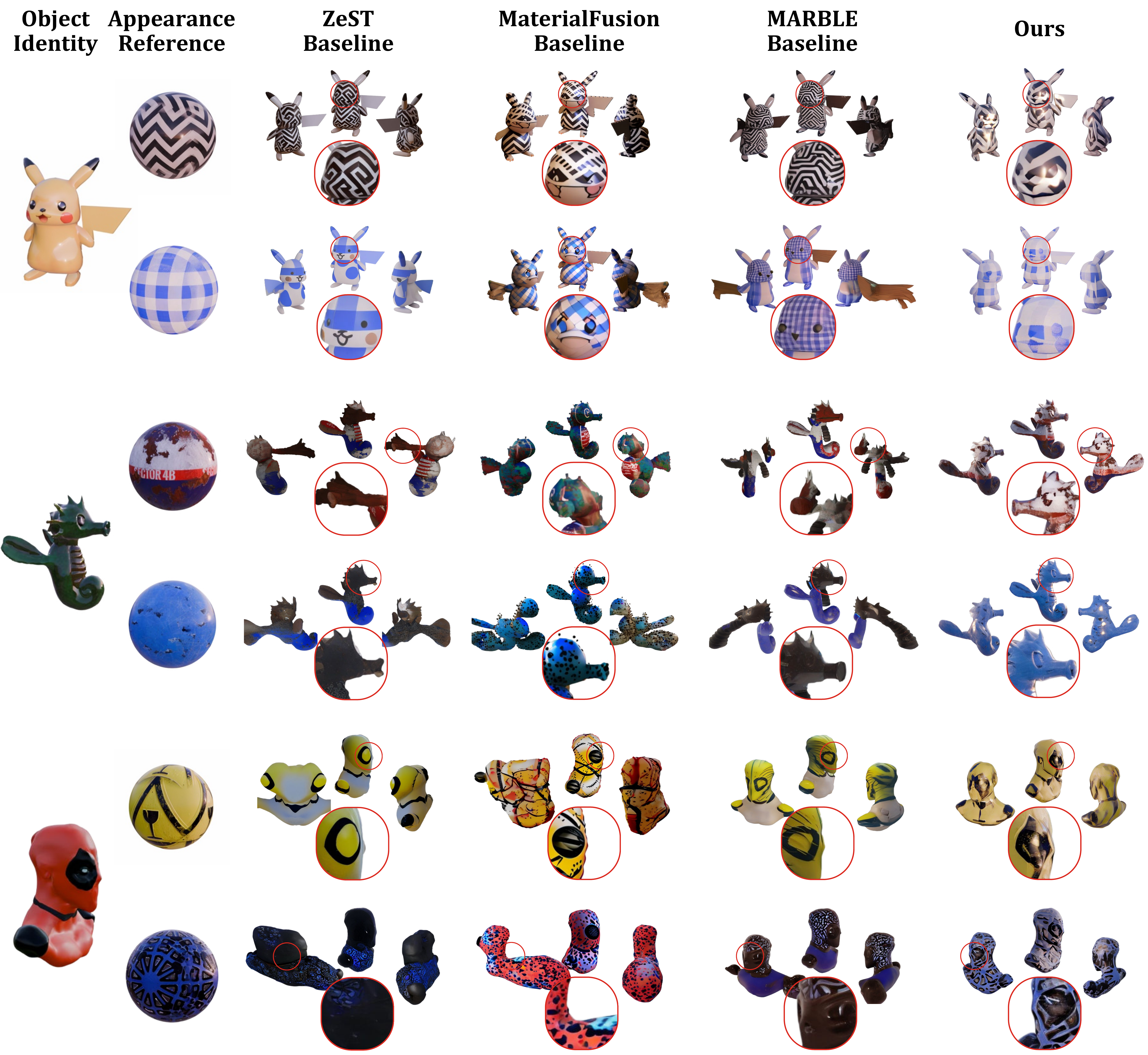}
    \caption{Additional visual results on rendered image. Our method successfully preserves object identity, keeps the details and achieves plausible material application.}
    \label{fig:rendered}
\end{figure}

\Cref{fig:ours_vs_baselines} presents a visual comparison of our method with selected baselines.
The results highlight a key challenge: applying a material to a single 2D image can degrade its structural identity, making the subsequent multi-view lifting process more difficult. For instance, some material-object combinations, such as chess piece made out of orange, are out-of-distribution for the base multiview diffusion model, leading to distorted or failed 3D generations by baseline methods (e.g. in row 15, even though MARBLE results are high quality, the multiview generation fails). Our approach, in contrast, better preserves the object's identity while faithfully applying the target material. Additional visual results of our method are presented in \Cref{fig:rendered}.

\begin{figure}[ht]
    \centering
    \includegraphics[width=\linewidth]{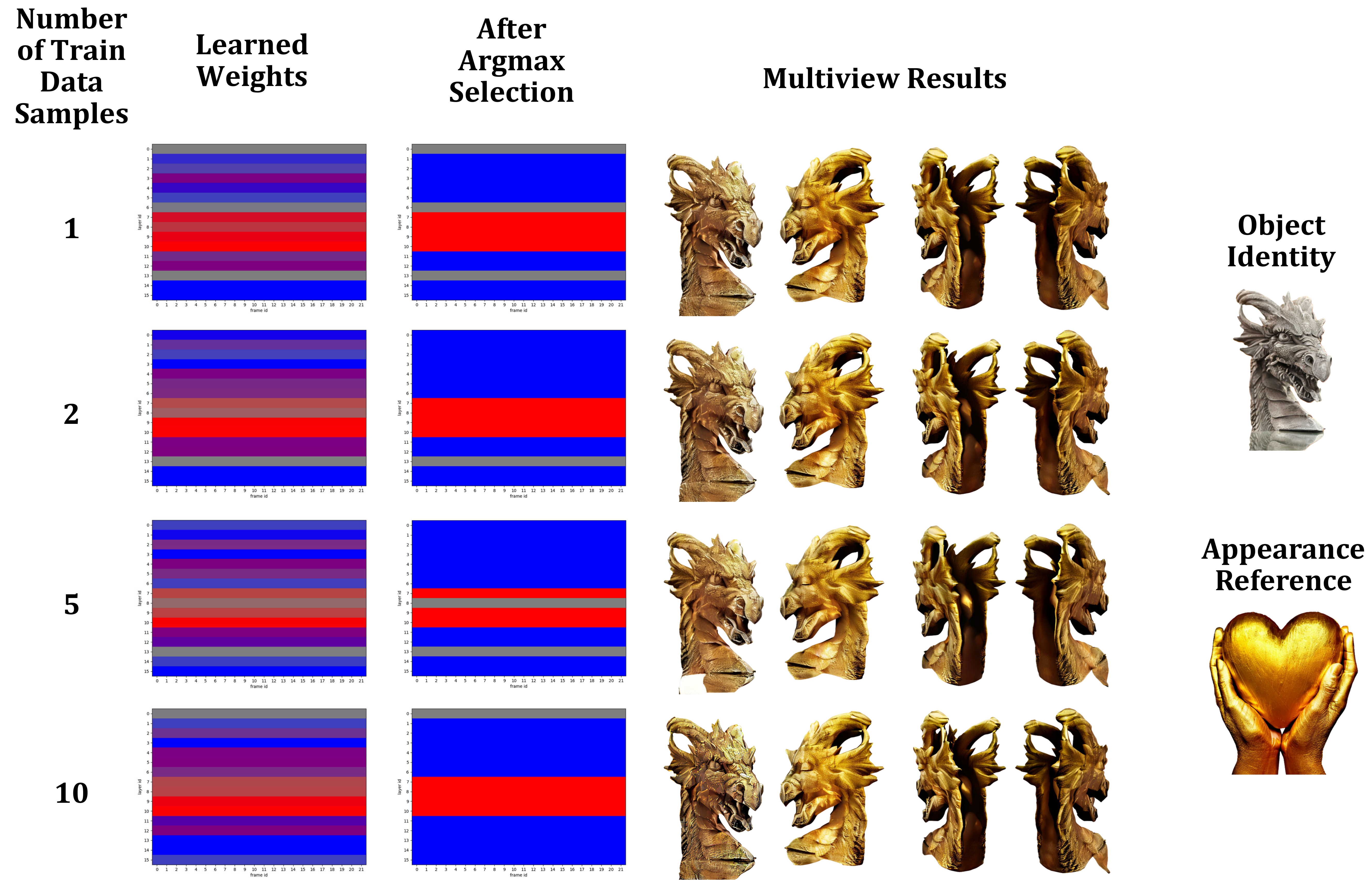}

    \caption{Data efficiency: our method needs as little as one training data sample to learn appearance control.}
    \label{fig:train_data}
\end{figure}

\noindent\textbf{Real-World, Generated and Rendered Images.}
To demonstrate the generalization ability of our approach, we apply it to both real-world photographs and synthetic images generated by diffusion models. As shown in \Cref{fig:ours_vs_baselines}, our method successfully transfers materials to real image inputs, as well as to generated images with complex textures and lighting. This highlights the robustness of our model beyond controlled renderings, enabling material manipulation on diverse and unconstrained visual inputs.

\noindent\textbf{Generalization to multiview materials.}
We can extend our reference appearance to multiview images so that multiview materials could be used during appearance transfer, enabling diverse materials for different object perspectives. The results are in the supplementary material. 

\noindent\textbf{Training set analysis.} We analyze the size of our training data set and its influence on the final results. We have observed that even one single pair of object and appearance reference is enough to optimize diffusion model with our method. The analysis of different train set lengths is specified in~\Cref{fig:train_data}.

\noindent\textbf{User study.}
To further assess perceptual performance, we conducted a users study involving 12 participants and 20 questions, each comparing our method against the three baselines.
For every query, participants were shown short video sequences generated by Ours, Marble, MaterialFusion, and ZeST on the same input.
They were asked to select the best method according to four criteria: detail preservation, material fidelity, multi-view consistency, and overall quality.
Across all categories, participants consistently preferred our method by a large margin.
Table~\ref{tab:userstudy} summarizes the vote distribution (240 total votes per category).

\begin{table}[t]
\centering
\caption{User study results. Each criterion contains 240 total votes. Participants consistently prefer our method across all four evaluation criteria.
}
\label{tab:userstudy}
\resizebox{\linewidth}{!}{
\begin{tabular}{lcccc}
\toprule
Criteria & \textbf{Ours} & MARBLE~\cite{cheng2025marble} & MaterialFusion~\cite{garifullin2025materialfusion} & ZeST~\cite{cheng2024zest} \\
\midrule
Detail Preservation 
& \textbf{154 (64.17\%) }
& 55 (22.92\%) 
& 20 (8.33\%) 
& 11 (4.58\%) \\
Material Fidelity 
& \textbf{171 (71.25\%)}
& 39 (16.25\%) 
& 14 (5.83\%) 
& 16 (6.67\%) \\
View Consistency 
& \textbf{140 (58.33\%) }
& 59 (24.58\%) 
& 23 (9.58\%) 
& 18 (7.50\%) \\
Overall Quality 
& \textbf{160 (66.67\%) }
& 49 (20.42\%) 
& 17 (7.08\%) 
& 14 (5.83\%) \\
\bottomrule
\end{tabular}
} 
\end{table}

\subsection{Ablation Study}
\label{subsec:ablation}

\begin{figure}[ht]
    \centering
    \includegraphics[width=\linewidth]{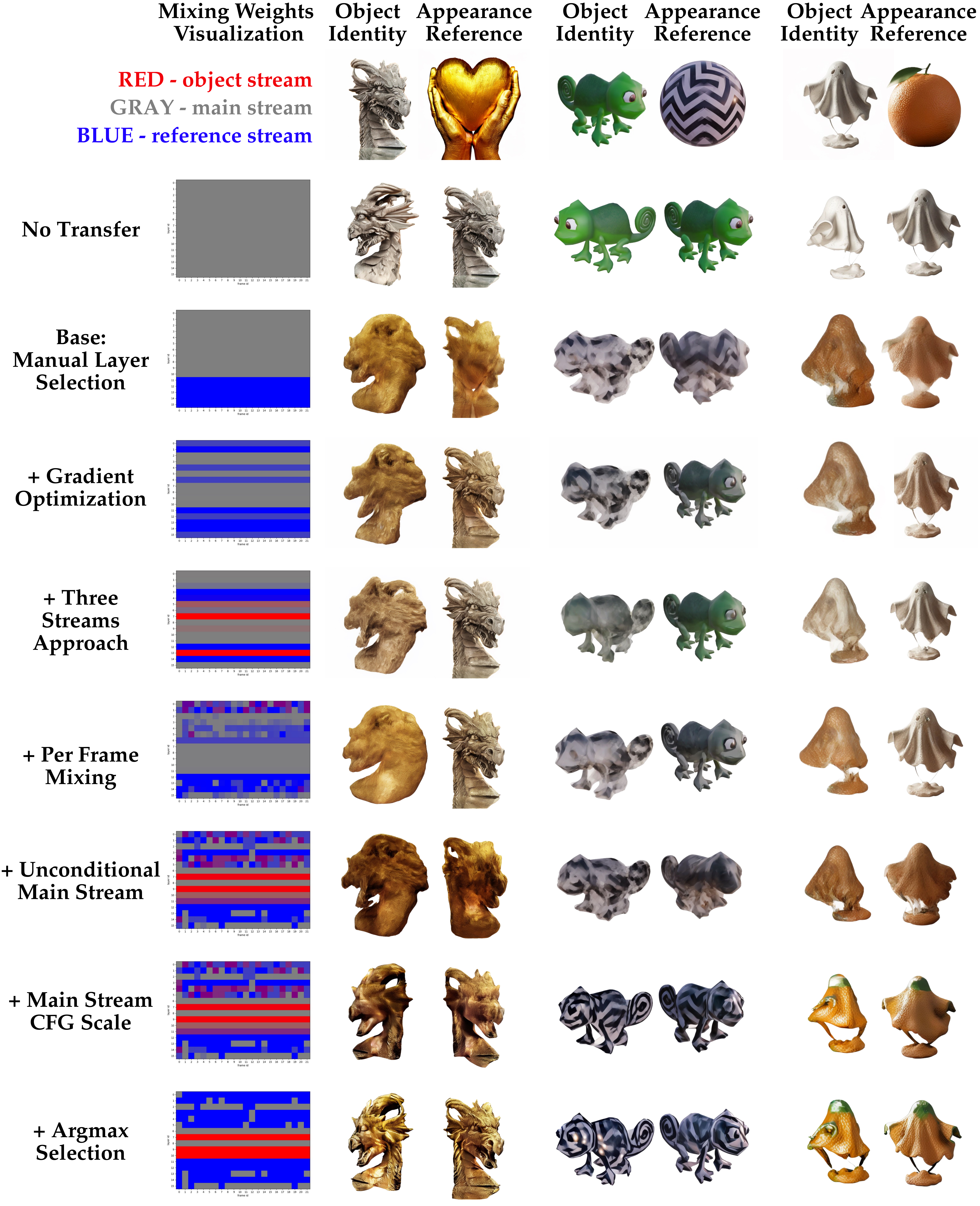}
    \caption{
        \textbf{Ablation study.}
        Each row shows the learned transfer weights (left) and three generated examples (right). A transfer weight map is plot with frame ID (0-21) on horizontal axis and layer ID on vertical axis. The weight magnitude is illustrated via the intensity of the red, grey, blue color. 
        Adding components progressively improves material fidelity, detail preservation, and multi-view consistency.
    }
    \label{fig:ablation}
\end{figure}

To evaluate the contribution of each component in our pipeline, we perform an ablation study shown in Fig.~\ref{fig:ablation}. 
Each row corresponds to one variant of our method, starting from the simplest 
\emph{Base (Manual Layer Selection)}, then progressively adding key design elements:  \emph{Gradient Optimization}, \emph{Three Streams Approach}, 
\emph{Unconditional Main Stream}, \emph{Per Frame Mixing}, \emph{Main Stream CFG Scale Adjustment}, and 
\emph{Argmax Selection}. 
For each variant, we visualize the learned mixing weights across self-attention layers (and later per frame), together with three representative output examples.

\noindent\textbf{Base: Manual Layer Selection.}
Following prior work, we manually select decoder self-attention layers to inject information from the reference stream.  
This introduces some reference cues but leads to blurry outputs and partial loss of object identity.  
Usually, selecting decoder-side layers works reasonably well, especially for style transfer, but the optimal choice always depends on the specific model.

\noindent\textbf{Gradient Optimization.}
Replacing manual selection with learnable mixing parameters optimized via gradients yields more effective transfers.  
The optimization often discovers a different and better set of layers than manual heuristics.

\noindent\textbf{Three Streams Approach.}
Our result confirms that introducing a third stream for object-identity images enables the model to preserve fine-grained geometry and structural cues, reducing identity drift in the generated results.

\noindent\textbf{Per Frame Mixing.}
Allowing mixing weights to vary per input frame increases flexibility, enabling the model to handle difficult viewpoints independently and improving robustness across views.

\noindent\textbf{Unconditional Main Stream.}
Removing the concatenation based conditioning from the main stream reduces its tendency to over-align with the input frame, enabling a more meaningful integration of shape and material cues from the other streams. We still preserve cross-attention (CLIP) conditioning in the Main Stream, using the mean CLIP embedding of the object and reference images.

\noindent\textbf{Main Stream CFG Scale Adjustment.}
Increasing the CFG scale of Main Stream (from 1.2–2.0 to 5.0–7.0) strengthens the influence of the appearance and identity streams, improving both global structure and material consistency.

\noindent\textbf{Argmax Selection.}
Instead of mixing streams with soft activations, we select the strongest stream per layer and per frame via an argmax rule during inference.  
This removes interference between sources, recovering sharper details and more stable appearance transfer.

%% file: sec/5_conclusion.tex
\section{Conclusion}
\label{sec:conclusion}

We presented a method for appearance transfer using self-attention transfer in image-to-multiview diffusion models. 
Our method offers a lightweight adaptation based on a simple design of the aggregator function that transfer object identity and reference appearance from input images into a final image. 
Our method can be applied for a variety of transfer tasks, including material, texture, lighting, and style transfer and only requires very few data samples for training. 
We showcase our method applied on SEVA~\cite{zhou2025seva} and Era3D~\cite{li2024era3d}, two recent state-of-the-art pretrained image-to-multiview diffusion models with favorable results. 

This research is not without limitations. 
Despite being lightweight, our method requires few-shot training, making it less accessible to users without GPU hardware and training data. 
Our pipeline remains general without any domain prior knowledge integrated, e.g., object standability, geometry complexity. 
Future research includes improvement on finegrained appearance control, extension to include text conditioning and controlling appearance in scenes. 

\paragraph{Acknowledgements}

Binh-Son Hua is supported by Research Ireland under the Research Ireland Frontiers for the Future Programme - Project, award number 22/FFP-P/11522.
This work was conducted with the financial support of the Research Ireland Centre for Research Training in Digitally Enhanced Reality (d-real) under Grant No. 18/CRT/6224. For the purpose of Open Access, the author has applied a CC BY public copyright licence to any Author Accepted Manuscript version arising from this submission.

%% file: sec/X_suppl.tex
\clearpage
\maketitlesupplementary

\begin{abstract}
In this supplementary package, we provide the following additional materials: 

    1) A video demonstrating a high-level overview of our method and our results, 

    2) Appearance transfer results on the Era3D pretrained model, 

    3) Appearance transfer based on multi-view references, i.e., our method can be easily extended to support different appearance references for object front, back, etc.,

    4) Simple appearance control by tweaking material parameters including roughness, metallic properties via reference images,

We also provide more discussions on our method including the training examples and current limitations.

\end{abstract}

\section{Multiview References}
Our method can take a multiview appearance reference which specifies how the object should look from different viewpoints. Concretely, we encode the multiview images into the latent space and, before every denoising step, replace the noisy latents in the appearance-reference stream with these clean multiview latents. For SEVA, we additionally mark all frames in the reference stream as source frames.

\begin{figure*}[ht]
    \centering
    \includegraphics[width=\linewidth]{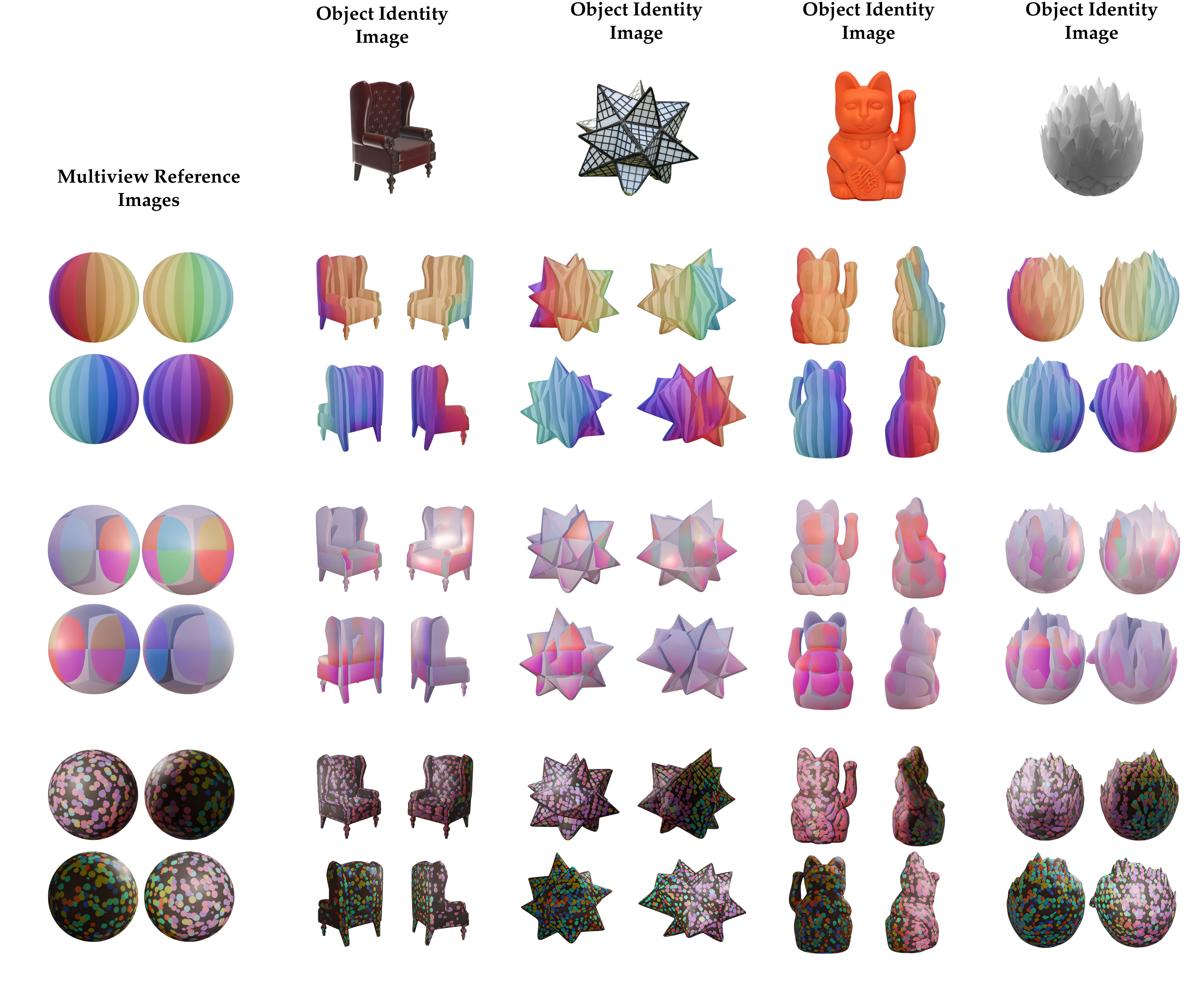}
    \caption{Visual results using a multiview appearance reference. FROMAT transfers view-dependent appearance consistently across viewpoints.}
    \label{fig:multiview}
\end{figure*}

The results are in~\Cref{fig:multiview}. 
As can be seen, FROMAT can output view-dependent material appearance from the sequence of reference images, and transfer them consistently to the target object across all views.

\section{Appearance Control}
\begin{figure*}[ht]
    \centering
    \includegraphics[width=\linewidth]{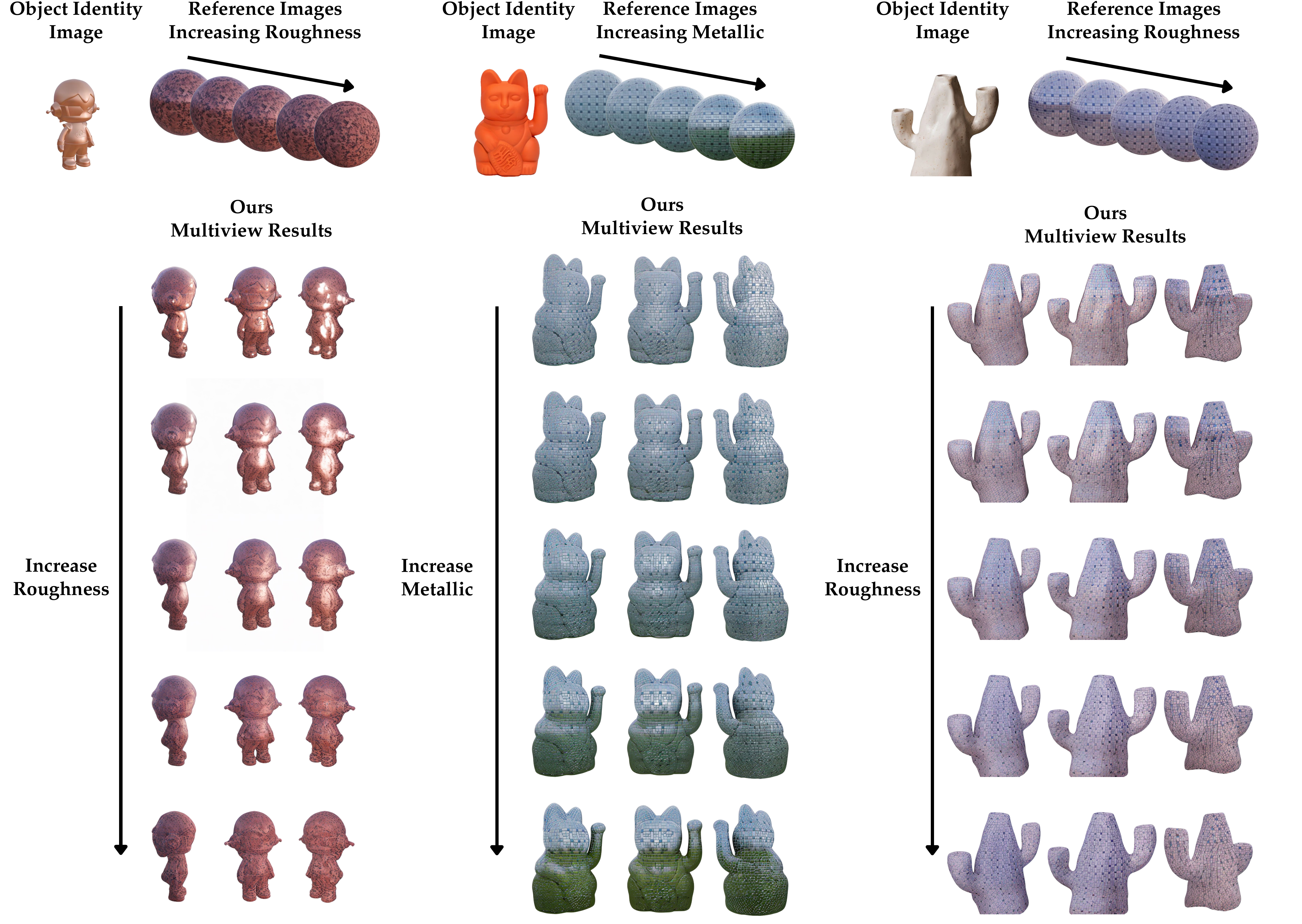}
    \caption{Control over material roughness and metalness via reference sphere exemplars.}
    \label{fig:precise}
\end{figure*}

\begin{figure*}[ht]
    \centering
    \includegraphics[width=\linewidth]{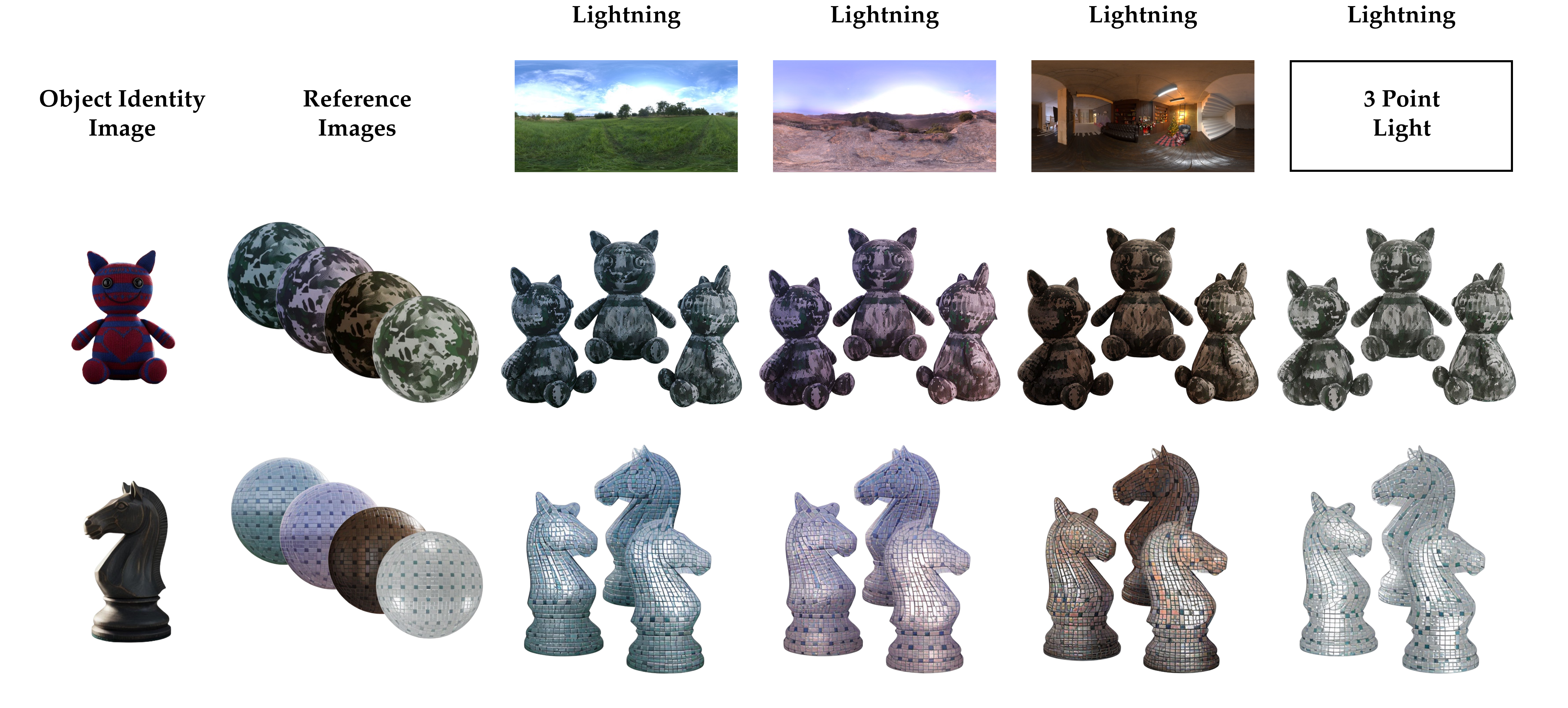}
    \caption{Control over different lighting conditions using rendered sphere references.}
    \label{fig:lighting}
\end{figure*}

Our method also supports appearance control over material properties and scene illumination. By rendering a reference sphere under different lighting and material settings, we can steer the lighting, metallic, and roughness of the transferred material (see~\Cref{fig:precise} and~\Cref{fig:lighting}). This demonstrates the potential for fine-grained appearance editing while preserving the underlying object identity and geometry.

\section{Support for Different Multiview Diffusion Models}
\begin{figure*}[ht]
    \centering
    \includegraphics[width=\linewidth]{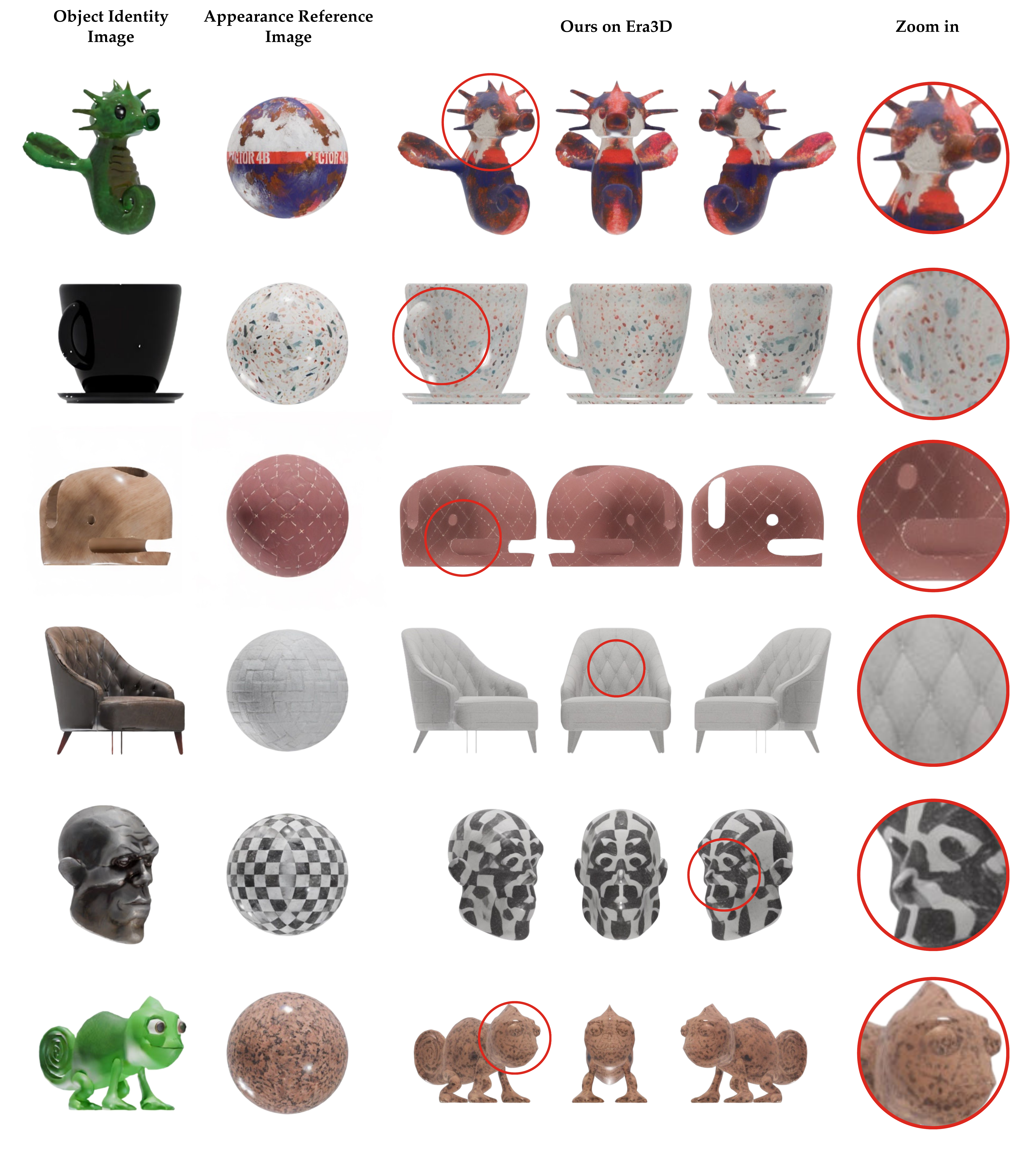}
    \caption{FROMAT applied to ERA3D: appearance transfer results.}
    \label{fig:results_era3d}
\end{figure*}

\begin{figure*}[ht]
    \centering
    \includegraphics[width=0.8\linewidth]{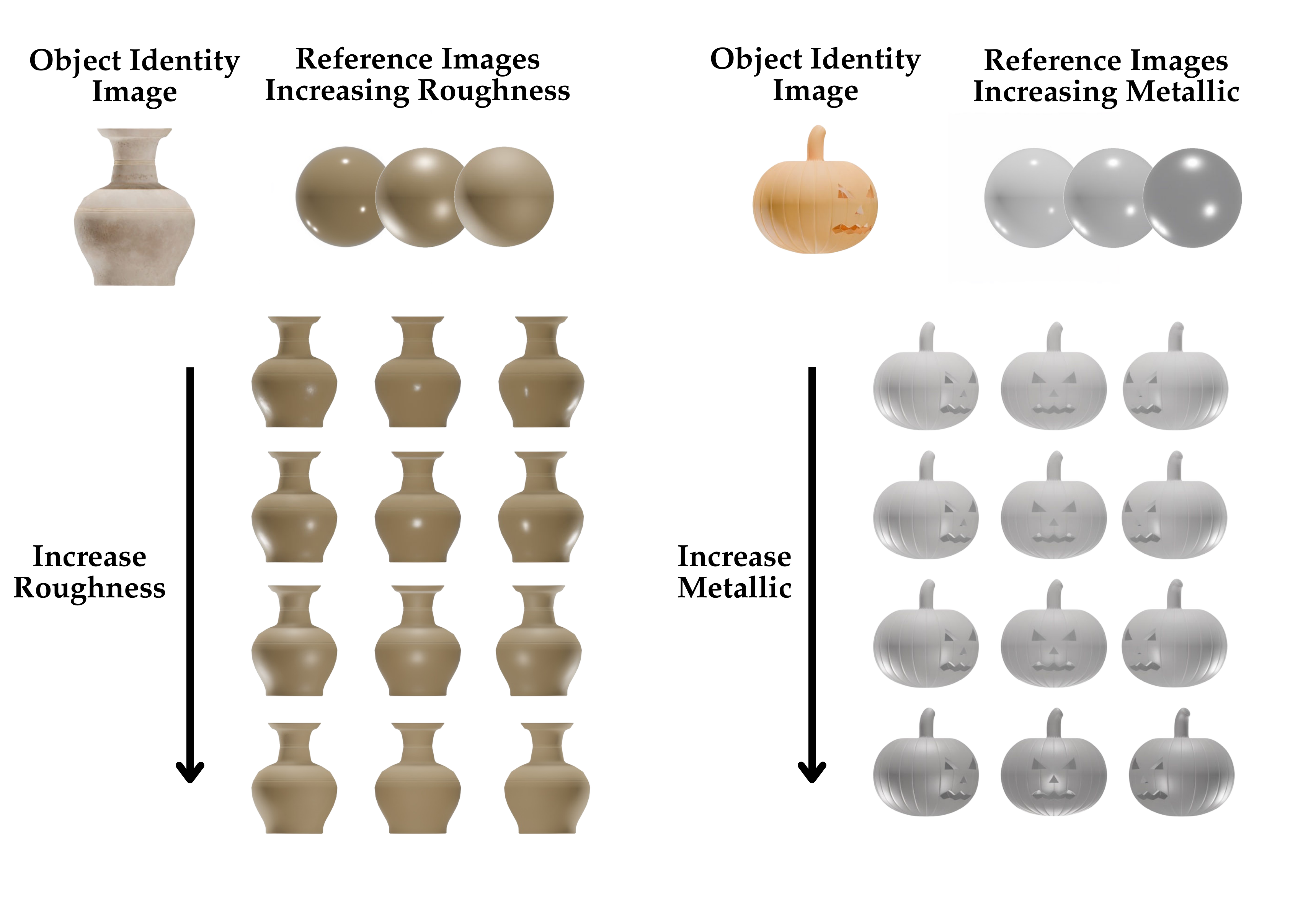}
    \caption{Control of material roughness and metalness on Era3D.}
    \label{fig:precise_era3d}
\end{figure*}

Our method is not restricted to Stable Virtual Camera~\cite{zhou2025seva}. In~\Cref{fig:results_era3d} and~\Cref{fig:precise_era3d}, we show results obtained with Era3D~\cite{li2024era3d}, demonstrating that FROMAT can be integrated with alternative multiview diffusion backbones. 
Compared to SEVA, Era3D outputs images at a fixed elevation angle and is limited to 6 distinct views along with 6 corresponding normal maps. Since Era3D is trained in an object-centric manner, we observe that its appearance transfer is quite detailed. However, we also find that it struggles on real or generated inputs, likely due to the restrictive rendering pipeline used during training.

Since our method operates on self-attention layers, we expect our method to be compatible with other multiview diffusion models that use self-attention transformer blocks. More explorations on other multiview models are left as future work.

\section{Camera Trajectories (SEVA)}
Although FROMAT is trained only on orbit-style camera trajectories, it can be applied to other camera paths provided by SEVA. In practice, we can reuse the camera control interface of the underlying Stable Virtual Camera model and still obtain consistent appearance transfer along more general trajectories. Please refer to the supplementary video for qualitative results on different camera trajectories.

\section{Guidelines for Choosing Training Examples}
To effectively demonstrate what should be transferred, training examples should emphasize three aspects: object identity, material fidelity, and multiview consistency. In practice, we recommend using examples with non-trivial geometry, complex materials (e.g., a mix of specular and diffuse components), and diverse textures (e.g., both patterned and solid-color regions). The examples should be challenging enough to exercise the capacity of the model, yet still sufficiently well-behaved so that the base multiview diffusion model can reliably reconstruct the unmodified object.

\begin{figure*}[ht]
    \centering
    \includegraphics[width=\linewidth]{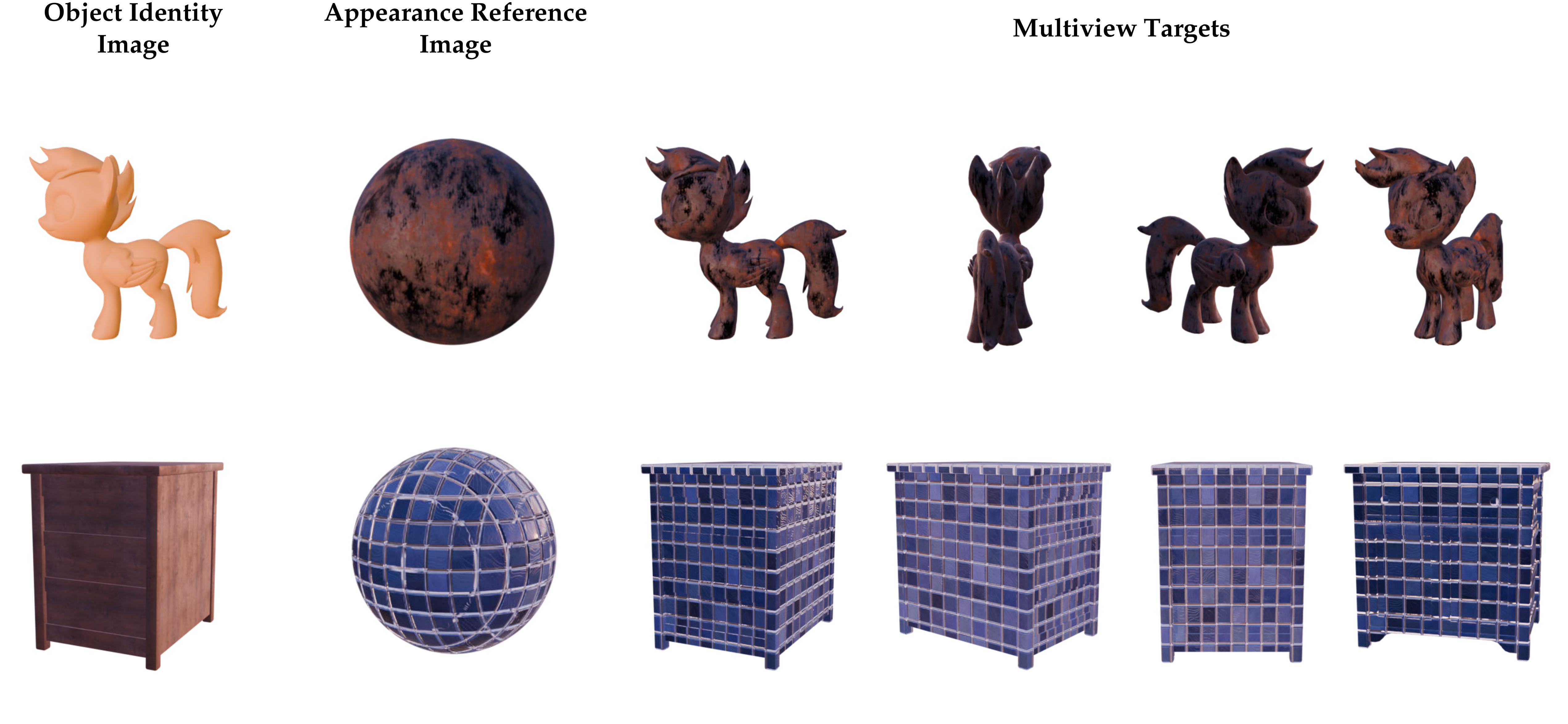}
    \caption{Training examples used for FROMAT. We select object–material pairs that balance challenging geometry and appearance with reliable reconstruction by the base multiview diffusion model.}
    \label{fig:train_set}
\end{figure*}

For our main results, we selected two object–material pairs, shown in~\Cref{fig:train_set}. We chose one object with complex geometry and predominantly diffuse materials, and one object with simpler geometry but specular materials and complex texture patterns.

\section{Limitations}
Currently, our method has the following limitations. First, we observe mild temporal flickering in some generated multiview sequences, especially under challenging lighting or high-frequency textures.
Second, in a few cases the model may remain too tightly aligned to the input object appearance, which restricts the magnitude of achievable appearance changes even though our input frame issue strategies alleviates this to some extent. Finally, extending FROMAT to manipulate appearance in scenes would require additional engineering on attention masking.

Despite these limitations, our method highlights the promise of multiview images as a powerful representation for implicit 3D generation and editing. We believe this line of research is still in its early stages, and there are many exciting opportunities for extensions and improvements in future work.

%% file: main.bib
@String(CVPR= {IEEE Conf. Comput. Vis. Pattern Recog.})

@String(ICCV= {Int. Conf. Comput. Vis.})

@String(ECCV= {Eur. Conf. Comput. Vis.})

@String(ICLR = {Int. Conf. Learn. Represent.})

@String(CVPR  = {CVPR})

@String(ICCV  = {ICCV})

@String(ECCV  = {ECCV})

@String(ICLR  = {ICLR})

@article{shi2024MVDream,
  author = {Shi, Yichun and Wang, Peng and Ye, Jianglong and Mai, Long and Li, Kejie and Yang, Xiao},
  title = {MVDream: Multi-view Diffusion for 3D Generation},
  journal = {ICLR},
  year = {2024},
}

@article{gao2024cat3d,
    title={CAT3D: Create Anything in 3D with Multi-View Diffusion Models},
    author={Ruiqi Gao* and Aleksander Holynski* and Philipp Henzler and Arthur Brussee and Ricardo Martin-Brualla and Pratul P. Srinivasan and Jonathan T. Barron and Ben Poole*
    },
    journal={Advances in Neural Information Processing Systems},
    year={2024}
}

@inproceedings{liu2024SyncDreamer,
      title={SyncDreamer: Generating Multiview-consistent Images from a Single-view Image}, 
      author={Yuan Liu and Cheng Lin and Zijiao Zeng and Xiaoxiao Long and Lingjie Liu and Taku Komura and Wenping Wang},
      year={2024},
    booktitle={ICLR}
}

@misc{zuo2024videomv,
      title={VideoMV: Consistent Multi-View Generation Based on Large Video Generative Model}, 
      author={Qi Zuo and Xiaodong Gu and Lingteng Qiu and Yuan Dong and Zhengyi Zhao and Weihao Yuan and Rui Peng and Siyu Zhu and Zilong Dong and Liefeng Bo and Qixing Huang},
      year={2024},
      eprint={2403.12010},
      archivePrefix={arXiv},
      primaryClass={cs.CV}
}

@misc{Wonder3D,
      title={Wonder3D: Single Image to 3D using Cross-Domain Diffusion}, 
      author={Xiaoxiao Long and Yuan-Chen Guo and Cheng Lin and Yuan Liu and Zhiyang Dou and Lingjie Liu and Yuexin Ma and Song-Hai Zhang and Marc Habermann and Christian Theobalt and Wenping Wang},
      year={2023},
      eprint={2310.15008},
      archivePrefix={arXiv},
      primaryClass={cs.CV},
      url={https://arxiv.org/abs/2310.15008}, 
}

@inproceedings{voleti2024sv3d,
  author    = {Voleti, Vikram and Yao, Chun-Han and Boss, Mark and Letts, Adam and Pankratz, David and Tochilkin,  Dmitrii and Laforte, Christian and Rombach, Robin and Jampani, Varun},
  title     = {{SV3D}: Novel Multi-view Synthesis and {3D} Generation from a Single Image using Latent Video Diffusion},
  booktitle = {ECCV},
  year      = {2024},
}

@article{huang2024materialanything,
  author = {Huang, Xin and Wang, Tengfei and Liu, Ziwei and Wang, Qing},
  title = {Material Anything: Generating Materials for Any 3D Object via Diffusion},
  journal = {CVPR},
  year = {2025}
}

@inproceedings{rombach2022high,
  title={High-resolution image synthesis with latent diffusion models},
  author={Rombach, Robin and Blattmann, Andreas and Lorenz, Dominik and Esser, Patrick and Ommer, Bj{\"o}rn},
  booktitle={CVPR},
  year={2022}
}

@article{ye2023ipadapter,
  title={IP-Adapter: Text Compatible Image Prompt Adapter for Text-to-Image Diffusion Models},
  author={Ye, Hu and Zhang, Jun and Liu, Sibo and Han, Xiao and Yang, Wei},
  journal={arXiv preprint arxiv:2308.06721},
  year={2023}
}

@article{huang2025mvadapter,
  title={MV-Adapter: Multi-view Consistent Image Generation Made Easy},
  author={Huang, Zehuan and Guo, Yuanchen and Wang, Haoran and Yi, Ran and Ma, Lizhuang and Cao, Yan-Pei and Sheng, Lu},
  journal={ICLR},
  year={2025}
}

@inproceedings{hertz2024stylealigned,
  title={Style Aligned Image Generation via Shared Attention},
  author={Hertz, Amir and Voynov, Andrey and Fruchter, Shlomi and Cohen-Or, Daniel},
  booktitle={CVPR},
  year={2024}
}

@article{jeong2024vsp,
      title={Visual Style Prompting with Swapping Self-Attention}, 
      author={Jaeseok Jeong and Junho Kim and Yunjey Choi and Gayoung Lee and Youngjung Uh},
      year={2024},
      journal={arXiv preprint arXiv:2402.12974},
}

@InProceedings{chung2024styleinjection,
    author    = {Chung, Jiwoo and Hyun, Sangeek and Heo, Jae-Pil},
    title     = {Style Injection in Diffusion: A Training-free Approach for Adapting Large-scale Diffusion Models for Style Transfer},
    booktitle = {CVPR},
    year      = {2024},
}

@inproceedings{jeong2025stylekeeper,
  title={StyleKeeper: Prevent Content Leakage using Negative Visual Query Guidance},
  author={Jeong, Jaeseok and Kim, Junho and Lee, Gayoung and Choi, Yunjey and Uh, Youngjung},
  booktitle={ICCV},
  year={2025}
}

@article{ruiz2022dreambooth,
  title={DreamBooth: Fine Tuning Text-to-image Diffusion Models for Subject-Driven Generation},
  author={Ruiz, Nataniel and Li, Yuanzhen and Jampani, Varun and Pritch, Yael and Rubinstein, Michael and Aberman, Kfir},
  journal={arXiv preprint arxiv:2208.12242},
  year={2022}
}

@inproceedings{sohn2023styledrop,
      title={StyleDrop: Text-to-Image Generation in Any Style}, 
      author={Kihyuk Sohn and Nataniel Ruiz and Kimin Lee and Daniel Castro Chin and Irina Blok and Huiwen Chang and Jarred Barber and Lu Jiang and Glenn Entis and Yuanzhen Li and Yuan Hao and Irfan Essa and Michael Rubinstein and Dilip Krishnan},
      year={2023},
    booktitle={NeurIPS}
}

@inproceedings{alaluf2024crossattn,
  title={Cross-Image Attention for Zero-Shot Appearance Transfer}, 
  author={Yuval Alaluf and Daniel Garibi and Or Patashnik and Hadar Averbuch-Elor and Daniel Cohen-Or},
  year={2024},
  booktitle={SIGGRAPH}
}

@article{wang2024instantstyle,
  title={InstantStyle-Plus: Style Transfer with Content-Preserving in Text-to-Image Generation},
  author={Wang, Haofan and Xing, Peng and Huang, Renyuan and Ai, Hao and Wang, Qixun and Bai, Xu},
  journal={arXiv preprint arXiv:2407.00788},
  year={2024}
}

@article{li2024era3d,   
  title={Era3D: High-Resolution Multiview Diffusion using Efficient Row-wise Attention},
  author={Li, Peng and Liu, Yuan and Long, Xiaoxiao and Zhang, Feihu and Lin, Cheng and Li, Mengfei and Qi, Xingqun and Zhang, Shanghang and Luo, Wenhan and Tan, Ping and others},
  journal={NeurIPS},
  year={2024}
}

@article{zhou2025seva,
      title={Stable Virtual Camera: Generative View Synthesis with Diffusion Models},
      author={Jensen (Jinghao) Zhou and Hang Gao and Vikram Voleti and Aaryaman Vasishta and Chun-Han Yao and Mark Boss and
      Philip Torr and Christian Rupprecht and Varun Jampani
      },
      journal={ICCV},
      year={2025}
  }

@article{cheng2024zest,
  title={ZeST: Zero-Shot Material Transfer from a Single Image},
  author={Cheng, Ta-Ying and Sharma, Prafull and Markham, Andrew and Trigoni, Niki and Jampani, Varun},
  journal={ECCV},
  year={2024}
}

@article{garifullin2025materialfusion,
  title={MaterialFusion: High-Quality, Zero-Shot, and Controllable Material Transfer with Diffusion Models},
  author={Kamil Garifullin and Maxim Nikolaev and Andrey Kuznetsov and Aibek Alanov},
  journal={CVPR},
 year={2025}
}

@article{cheng2025marble,
  title={MARBLE: Material Recomposition and Blending in CLIP-Space},
  author={Cheng, Ta-Ying and Sharma, Prafull and Boss, Mark and Jampani, Varun},
  journal={CVPR},
  year={2025}
}

@InProceedings{gar,
author="Titov, Vadim
and Khalmatova, Madina
and Ivanova, Alexandra
and Vetrov, Dmitry
and Alanov, Aibek",
title="Guide-and-Rescale: Self-guidance Mechanism for~Effective Tuning-Free Real Image Editing",
booktitle="ECCV",
year="2024",
}

@misc{Fortin2025NanoBanana,
  author       = {Alisa Fortin and Guillaume Vernade and Kat Kampf and Ammaar Reshi},
  title        = {Introducing {Gemini} 2.5 Flash Image, our state‑of‑the‑art image model (aka “nano‑banana”)},
  howpublished = {Google Developers Blog},
  month        = aug,
  year         = {2025},
  note         = {Available at: \url{https://developers.googleblog.com/en/introducing-gemini-2-5-flash-image/} (accessed Nov. 2025)}
}

@inproceedings{haque2023instructnerf,
    author = {Haque, Ayaan and Tancik, Matthew and Efros, Alexei and Holynski, Aleksander and Kanazawa, Angjoo},
    title = {Instruct-NeRF2NeRF: Editing 3D Scenes with Instructions},
    booktitle = {ICCV},
    year = {2023},
}

@InProceedings{shum2024fusion,
    author    = {Shum, Ka Chun and Kim, Jaeyeon and Hua, Binh-Son and Nguyen, Duc Thanh and Yeung, Sai-Kit},
    title     = {Language-driven Object Fusion into Neural Radiance Fields with Pose-Conditioned Dataset Updates},
    booktitle = {CVPR},
    year      = {2024},
}

@inproceedings{engelhardt2025svim3d, author ={Engelhardt, Andreas and Boss, Mark and Voleti, Vikram and Yao, Chun-Han and Lensch, Hendrik P.A. and Jampani, Varun},title ={{SViM3D}: Stable Video Material Diffusion for Single Image 3D Generation},booktitle ={ICCV},year ={2025}}

@inproceedings{lopes2024materialpalette,
    author = {Lopes, Ivan and Pizzati, Fabio and de Charette, Raoul},
    title = {Material Palette: Extraction of Materials from a Single Image},
    booktitle = {CVPR},
    year = {2024},
}

@inproceedings{sharma2024alchemist,
      title={Alchemist: Parametric Control of Material Properties with Diffusion Models}, 
      author={Prafull Sharma and Varun Jampani and Yuanzhen Li and Xuhui Jia and Dmitry Lagun and Fredo Durand and William T. Freeman and Mark Matthews},
      year={2024},
    booktitle={CVPR}
}

@article{vainer2024pbr,
  author    = {Vainer, Shimon and Boss, Mark and Parger, Mathias and Kutsy, Konstantin and 
               De Nigris, Dante and Rowles, Ciara and Perony, Nicolas and Donn\'e, Simon},
  title     = {Collaborative Control for Geometry-Conditioned PBR Image Generation},
  journal   = {arXiv preprint arXiv:2402.05919},
  year      = {2024},
}

@inproceedings{deitke2023objaverse,
  title={Objaverse: A Universe of Annotated 3D Objects},
  author={Matt Deitke and Dustin Schwenk and Jordi Salvador and Luca Weihs and
          Oscar Michel and Eli VanderBilt and Ludwig Schmidt and
          Kiana Ehsani and Aniruddha Kembhavi and Ali Farhadi},
  booktitle={CVPR},
  year={2023}
}

@inproceedings{vecchio2024matsynth,
  author    = {Vecchio, Giuseppe and Deschaintre, Valentin},
  title     = {MatSynth: A Modern PBR Materials Dataset},
  booktitle = {CVPR},
  year      = {2024},
}

@article{vecchio2024controlmat,
  author = {Vecchio, Giuseppe and Martin, Rosalie and Roullier, Arthur and Kaiser, Adrien and Rouffet, Romain and Deschaintre, Valentin and Boubekeur, Tamy},
  title = {ControlMat: A Controlled Generative Approach to Material Capture},
  year = {2024},
  journal = {ACM Trans. Graph.},
}

@article{guo2020MaterialGAN,
  title={MaterialGAN: Reflectance Capture using a Generative SVBRDF Model},
  author={Guo, Yu and Smith, Cameron and Ha\v{s}an, Milo\v{s} and Sunkavalli, Kalyan and Zhao, Shuang},
  journal={ACM Trans. Graph.},
  year={2020},
}

@article{hu2022material,
author = {Hu, Yiwei and Hašan, Miloš and Guerrero, Paul and Rushmeier, Holly and Deschaintre, Valentin},
title = {Controlling Material Appearance by Examples},
journal = {Computer Graphics Forum},
year = {2022}
}

@inproceedings{zhou2022tilegen,
  title={TileGen: Tileable, Controllable Material Generation and Capture},
  author={Zhou, Xilong and Hasan, Milos and Deschaintre, Valentin and Guerrero, Paul and Sunkavalli, Kalyan and Kalantari, Nima Khademi},
  booktitle={SIGGRAPH Asia 2022 Conference Papers},
  year={2022}
}

@article{ma2025picker,
author = {Ma, Xiaohe and Deschaintre, Valentin and Ha\v{s}an, Milo\v{s} and Luan, Fujun and Zhou, Kun and Wu, Hongzhi and Hu, Yiwei},
title = {MaterialPicker: Multi-Modal DiT-Based Material Generation},
year = {2025},
journal = {ACM Trans. Graph.},
}

@inproceedings{zeng2024rgb,
author = {Zeng, Zheng and Deschaintre, Valentin and Georgiev, Iliyan and Hold-Geoffroy, Yannick and Hu, Yiwei and Luan, Fujun and Yan, Ling-Qi and Ha\v{s}an, Milo\v{s}},
title = {RGB-X: Image decomposition and synthesis using material- and lighting-aware diffusion models},
year = {2024},
booktitle = {ACM SIGGRAPH 2024 Conference Papers},
}

@article{lyu2025intrinsicedit,
author = {Lyu, Linjie and Deschaintre, Valentin and Hold-Geoffroy, Yannick and Ha\v{s}an, Milo\v{s} and Yoon, Jae Shin and Leimk\"{u}hler, Thomas and Theobalt, Christian and Georgiev, Iliyan},
title = {IntrinsicEdit: Precise generative image manipulation in intrinsic space},
year = {2025},
journal = {ACM Trans. Graph.},
}

@article{lopes2025matswap,
author = {Lopes, I. and Deschaintre, V. and Hold-Geoffroy, Y. and de Charette, R.},
title = {MatSwap: Light-aware material transfers in images},
journal = {Computer Graphics Forum},
year = {2025}
}

@InProceedings{brooks2022instructpix2pix,
    author     = {Brooks, Tim and Holynski, Aleksander and Efros, Alexei A.},
    title      = {InstructPix2Pix: Learning to Follow Image Editing Instructions},
    booktitle  = {CVPR},
    year       = {2023},
}

@InProceedings{zhang2023controlnet,
    author    = {Zhang, Lvmin and Rao, Anyi and Agrawala, Maneesh},
    title     = {Adding Conditional Control to Text-to-Image Diffusion Models},
    booktitle = {ICCV},
    year      = {2023},
}

@inproceedings{mildenhall2020nerf,
  title={Nerf: Representing scenes as neural radiance fields for view synthesis},
  author={Mildenhall, Ben and Srinivasan, Pratul P and Tancik, Matthew and Barron, Jonathan T and Ramamoorthi, Ravi and Ng, Ren},
  booktitle={ECCV},
  year={2020},
}

@Article{kerbl3Dgaussians,
      author       = {Kerbl, Bernhard and Kopanas, Georgios and Leimk{\"u}hler, Thomas and Drettakis, George},
      title        = {3D Gaussian Splatting for Real-Time Radiance Field Rendering},
      journal      = {ACM Transactions on Graphics},
      year         = {2023},
}

@inproceedings{pang2023locally,
      title={Locally Stylized Neural Radiance Fields}, 
      author={Hong-Wing Pang and Binh-Son Hua and Sai-Kit Yeung},
      year={2023},
    booktitle={ICCV}, 
}

@article{liu2024stylegaussian,
  title={StyleGaussian: Instant 3D Style Transfer with Gaussian Splatting},
  author={Liu, Kunhao and Zhan, Fangneng and Xu, Muyu and Theobalt, Christian and Shao, Ling and Lu, Shijian},
  journal={SIGGRAPH Asia 2024 Technical Communications},
  year={2024},
}

@article{zhang2024dreammat,
author = {Zhang, Yuqing and Liu, Yuan and Xie, Zhiyu and Yang, Lei and Liu, Zhongyuan and Yang, Mengzhou and Zhang, Runze and Kou, Qilong and Lin, Cheng and Wang, Wenping and Jin, Xiaogang},
title = {DreamMat: High-quality PBR Material Generation with Geometry- and Light-aware Diffusion Models},
year = {2024},
journal = {ACM Trans. Graph.}, 
}

@article{verbin2022refnerf,
    title={{Ref-NeRF}: Structured View-Dependent Appearance for
           Neural Radiance Fields},
    author={Dor Verbin and Peter Hedman and Ben Mildenhall and
            Todd Zickler and Jonathan T. Barron and Pratul P. Srinivasan},
    journal={CVPR},
    year={2022}
}

@article{zhang2025refgs,
  title={Ref-GS: Directional Factorization for 2D Gaussian Splatting},
  author={Zhang, Youjia and Chen, Anpei and Wan, Yumin and Song, Zikai and Yu, Junqing and Luo, Yawei and Yang, Wei},
  journal={CVPR},
  year={2025}
}

@inproceedings{zhang2025materialrefgs,
  title={{MaterialRefGS}: Reflective Gaussian Splatting with Multi-view Consistent Material Inference},
  author={Zhang, Wenyuan and Tang, Jimin and Zhang, Weiqi and Fang, Yi and Liu, Yu-Shen and Han, Zhizhong},
  booktitle={Advances in Neural Information Processing Systems},
  year={2025}
}

@inproceedings{ye2024gsdr,
  author    = {Keyang, Ye and Qiming, Hou and Kun, Zhou},
  title     = {3D Gaussian Splatting with Deferred Reflection},
  booktitle  = {SIGGRAPH Conference},
  year      = {2024},
}

@article{jiang2024gaussianshader,
  title={GaussianShader: 3D Gaussian Splatting with Shading Functions for Reflective Surfaces},
  author={Jiang, Yingwenqi and Tu, Jiadong and Liu, Yuan and Gao, Xifeng and Long, Xiaoxiao and Wang, Wenping and Ma, Yuexin},
  journal={CVPR},
  year={2024}
}

@inproceedings{verbin2024nerfcasting,
author = {Verbin, Dor and Srinivasan, Pratul P. and Hedman, Peter and Mildenhall, Ben and Attal, Benjamin and Szeliski, Richard and Barron, Jonathan T.},
title = {NeRF-Casting: Improved View-Dependent Appearance with Consistent Reflections},
year = {2024},
booktitle = {SIGGRAPH Asia 2024 Conference Papers},
}

@inproceedings{kawar2023imagic,
      title={Imagic: Text-Based Real Image Editing with Diffusion Models},
      author={Kawar, Bahjat and Zada, Shiran and Lang, Oran and Tov, Omer and Chang, Huiwen and Dekel, Tali and Mosseri, Inbar and Irani, Michal},
      booktitle={CVPR},
      year={2023}
}

@misc{xie2024sana,
      title={Sana: Efficient High-Resolution Image Synthesis with Linear Diffusion Transformer},
      author={Enze Xie and Junsong Chen and Junyu Chen and Han Cai and Haotian Tang and Yujun Lin and Zhekai Zhang and Muyang Li and Ligeng Zhu and Yao Lu and Song Han},
      year={2024},
      eprint={2410.10629},
      archivePrefix={arXiv},
      primaryClass={cs.CV},
      url={https://arxiv.org/abs/2410.10629},
    }

@misc{xie2025sana,
      title={SANA 1.5: Efficient Scaling of Training-Time and Inference-Time Compute in Linear Diffusion Transformer},
      author={Xie, Enze and Chen, Junsong and Zhao, Yuyang and Yu, Jincheng and Zhu, Ligeng and Lin, Yujun and Zhang, Zhekai and Li, Muyang and Chen, Junyu and Cai, Han and others},
      year={2025},
      eprint={2501.18427},
      archivePrefix={arXiv},
      primaryClass={cs.CV},
      url={https://arxiv.org/abs/2501.18427},
    }

@inproceedings{poole2022dreamfusion,
title={DreamFusion: Text-to-3D using 2D Diffusion},
author={Ben Poole and Ajay Jain and Jonathan T. Barron and Ben Mildenhall},
booktitle={ICLR},
year={2023}
}

@InProceedings{chen2023fantasia3d,
  author={Chen, Rui and Chen, Yongwei and Jiao, Ningxin and Jia, Kui},
  title={Fantasia3D: Disentangling Geometry and Appearance for High-quality Text-to-3D Content Creation},
  booktitle={ICCV},
  year={2023},
}

@article{chen2023text2tex,
    title={Text2Tex: Text-driven Texture Synthesis via Diffusion Models},
    author={Chen, Dave Zhenyu and Siddiqui, Yawar and Lee, Hsin-Ying and Tulyakov, Sergey and Nie{\ss}ner, Matthias},
    journal={ICCV},
    year={2023},
}

@inproceedings{youwang2024paintit,
    title = {Paint-it: Text-to-Texture Synthesis via Deep Convolutional Texture Map Optimization and Physically-Based Rendering},
    author = {Youwang, Kim and Oh, Tae-Hyun and Pons-Moll, Gerard},
    booktitle = {CVPR},
    year = {2024}
}

@article{huo2024texgen,
    author    = {Huo, Dong and Guo, Zixin and Zuo, Xinxin and Shi, Zhihao and Lu, Juwei and Dai, Peng and Xu, Songcen and Cheng, Li and Yang, Yee-Hong},
    title     = {TexGen: Text-Guided 3D Texture Generation with Multi-view Sampling and Resampling},
    journal   = {ECCV},
    year      = {2024},
}

@inproceedings{deng2024flashtex,
  title={FlashTex: Fast Relightable Mesh Texturing with LightControlNet},
  author={Deng, Kangle and Omernick, Timothy and Weiss, Alexander and Ramanan, Deva and Zhu, Jun-Yan and Zhou, Tinghui and Agrawala, Maneesh},
  booktitle={ECCV},
  year={2024},
}

@article{chen2025advances,
  title={Advances in 3D neural stylization: a survey},
  author={Chen, Yingshu and Shao, Guocheng and Shum, Ka Chun and Hua, Binh-Son and Yeung, Sai-Kit},
  journal={International Journal of Computer Vision},
  pages={1--36},
  year={2025},
  publisher={Springer}
}
